\documentclass[conference]{IEEEtran}
\IEEEoverridecommandlockouts
\usepackage{cite}
\usepackage{amsmath,amssymb,amsfonts}
\usepackage{algorithmic}
\usepackage{graphicx}
\usepackage{textcomp}
\usepackage{xcolor}

\def\BibTeX{{\rm B\kern-.05em{\sc i\kern-.025em b}\kern-.08em
    T\kern-.1667em\lower.7ex\hbox{E}\kern-.125emX}}
\begin{document}

\title{Learning Enhancement of CNNs via Separation Index Maximizing at the First Convolutional Layer
}

\makeatletter
\newcommand{\linebreakand}{%
  \end{@IEEEauthorhalign}
  \hfill\mbox{}\par
  \mbox{}\hfill\begin{@IEEEauthorhalign}
}
\makeatother

\author{\IEEEauthorblockN{Ali Karimi}
\IEEEauthorblockA{\textit{School of Electrical and Computer Engineering} \\
\textit{College of Engineering, University Of Tehran}\\
Tehran, Iran \\
aliiikarimi@ut.ac.ir}
\and
\IEEEauthorblockN{Ahmad Kalhor}
\IEEEauthorblockA{\textit{School of Electrical and Computer Engineering} \\
\textit{College of Engineering, University Of Tehran}\\
Tehran, Iran \\
akalhor@ut.ac.ir} 
}

\maketitle

\begin{abstract}
In this paper, a straightforward enhancement learning algorithm based on Separation Index (SI) concept is proposed for Convolutional Neural Networks (CNNs). At first, the SI as a supervised complexity measure is explained its usage in better learning of CNNs for classification problems illustrate. Then, a learning strategy proposes through which the first layer of a CNN is optimized by maximizing the SI, and the further layers are trained through the backpropagation algorithm to learn further layers. In order to maximize the SI at the first layer, A variant of ranking loss is optimized by using the quasi least square error technique. Applying such a learning strategy to some known CNNs and datasets, its enhancement impact in almost all cases is demonstrated. 
\end{abstract}


\section{Introduction}

In recent years, deep learning has made significant progress. deep learning has been used in various fields, including video, image, audio, and data analysis. Extensive applications of deep learning and their ability to solve various problems have led to various studies to improve their performance, research on improving neural network training methods, layers used in the neural network, data preprocessing. The accuracy of deep learning-based classification algorithms is still one of the challenges in this field. So far, architectures such as VGG \cite{simonyan2014very}, AlexNet \cite{krizhevsky2009learning}, ResNet \cite{he2016deep}, InceptionV3 \cite{szegedy2016rethinking} based on learning with loss function and backpropagation method have been presented that have been able to provide good accuracy in Reach classification but still have much potential for improvement.

In this study, by changing the method of neural network training from a completely backward method to a combined method of feedforward and backward, in addition to reducing network training time, with better generalization, network accuracy in classification is also increased. In the proposed method, the first layer of the network is trained as feedforward, then the output of the first layer is given as input to the second layer, and the network is trained backward with the Adam optimizer.

The feedfowrd training method is that first a batch normalization layer is applied to the data, then the input data patches are extracted and a pincipal component analysis (PCA) is applied to the data using these patches. The eigenvalues obtained by PCA in the convolution layer are used as the values of the filters. Next, the output of the convolution layer is received and Quasi Least Square (QLS) is applied on them, which is used to calculate the new values of the convolution layer filters. This continues until we reach the best classification accuracy in the first layer.

In the following, Sec. Two reviews previous works, Sec. Three  describes our implementation details, and Sec. Four presents our experimental results. Finally, Sec. Five concludes the paper.

\section{Related Works}

In recent years, many limitations and challenges of deep learning algorithms have been addressed, however, there are a number of different shortcomings in learning methods. Initialization, network regularization, whitening, and choosing efficient loss functions are the main techniques for solving the complexity of deep neural network training that remain open to studying neural networks. In doing so, we adopt consistent regularization and initialization approaches to get a higher level of generalization, robustness and learning performance than other traditional types.

\begin{table*}[h]
\centering

\begin{center}
\begin{tabular}{|c|c|}
\hline
\textbf{Complexity measures} & \textbf{Overall evaluating approach}  \\
\hline\hline
Feature-based  & Discovering informative features by evaluating each feature independently \\
&  \cite{orriols2010documentation,cummins2013combining}) \\
Linearity separation & Evaluating the linearly separation of  different classes \\
&   \cite{bottou2007support} \\
Neighborhood  & Evaluating the shape of the decision boundary to distinguish different classes overlap \\
&   \cite{lorena2012analysis,leyva2014set} \\
Network  & Evaluating the data dataset structure and relationships by representing it as a graph \\
&   \cite{garcia2015effect} \\
Dimensionality  & Evaluating the sparsity of the data and the average number of features at each \\
&   dimension \cite{lorena2012analysis,basu2006data} \\
Class imbalanced  & Evaluating the proportion of dataset number between different classes \\
&   \cite{lorena2012analysis} \\
\hline
\end{tabular}
\end{center}
\caption{Some complexity measures and their evaluating approaches in a classification problem}
\end{table*}

\subsection{Parameter initialization}

The research of parameter initialization categorizes into three categories: 1- random initialization, 2- layer-wise initialization, and 3- decomposition-based initialization. Random initialization relies on setting up parameters haphazardly using a random-number generator from a specific distribution considering the network structural characteristics. The two popular random initializations by He et al. \cite{he2015delving} and Glorot et al. \cite{glorot2010understanding} set the layer weights in a preserving way of the input variance. Hinton et al. \cite{hinton2006reducing} and Bengio et al. \cite{bengio2007greedy} have proposed greedy layer-wise unsupervised learning policies. Saxe et al. \cite{ saxe2013exact}, Sussillo et al.  Wagner et al. \cite{wagner2013learning} follow a patch-based scheme to calculate the eigenvectors of the layer’s input and assign each kernel weight with an eigenvector in order of their corresponding eigenvalues. \cite{ sussillo2014random}, Mishkin et al. \cite{mishkin2015all} present an algorithmic layer-wise method that possesses two main steps.

At the pre-initialize step, network layers are firstly initialized with Gaussian noise and then replaced these weights with the components of its orthonormal biases.
Moreover, Krähenbühl et al. \cite{ krahenbuhl2015data} present other common random initializations In which random orthogonal initialization, adjusting weights via a scaling factor to keep the input/output error rate around one and data-driven channel-wise scaling mechanisms be used, respectively. This method, after pre-initializing, performs a layer by layer initialization in a data-driven manner scales its layer weights by the standard deviation of a mini-batch to keep the input variance. Wang et al. \cite{wang2020pca} introduce another PCA-based initialization policy in which it calculates the principal components for all layers in two stages. While the first step, the eigenvector matrix of the input patch covariance is determined; in the next step, the second eigenvector matrix is computed on the input’s projection by the first eigenvector matrix. In the end, this method allocates one of the matrices achieved from computing two eigenvectors’ outer production to each kernel window.

\subsection{Separation index as a complexity measure}

To evaluate the challenges of the dataset in a classification problem, several complexity measures have been introduced, \cite{kalhor2019evaluation}. Table 1 presents some introduced complexity measures and their overall evaluating approaches.

According to the given explanations about the Separation index in \cite{kalhor2019evaluation}, SI indicates that how much the data points with different labels are separated from each other. In addition, it has explained that while the SI increases layer by layer in a deep neural network (DNN), the margin among different classes will increase and the generalization of the network increases.  On the other hand, increasing the margin between different classes leads to boundary complexity reduction and hence one can consider SI as a variant of neighborhood measures. Fig \ref{fig:1} shows an illustrative example where data points of three classes (with the rectangle, circle and triangle indicators) have (a) low, (b) medium, or (c) high separation indices. As it is seen, when the SI increases in (c), the complexity boundaries between data points of three classes are less than the cases (a) and (b).  

\begin{figure*}[h!]
\includegraphics[width=\linewidth]{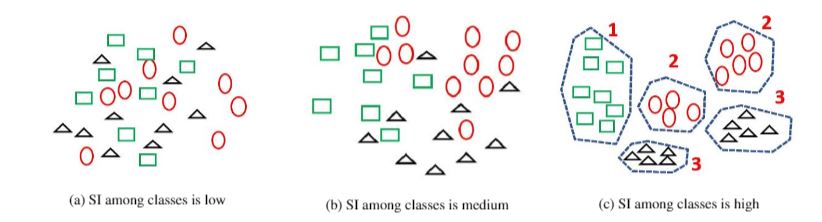}
\caption{An illustrative example where data points of three classes (with the rectangle, circle, and triangle indicators) have (a) low, (b) medium, or (c) high separation indices.}
\label{fig:1}
\end{figure*}

The first order SI, in fact counts the number of all data points having the same labels with their nearest data points (neighbors with minimum distances), \cite{kalhor2019evaluation}: 

\begin{equation}
\begin{array}{c}
\operatorname{SI}\left(\left\{\mathbf{z}^{\mathrm{q}}\right\}_{q=1}^{Q},\left\{l^{q}\right\}_{q=1}^{Q}\right)=\frac{1}{Q} \sum_{k=1}^{Q} \varphi\left(l^{q}-l^{q_{n e a r}}\right) \\ \\
\varphi(v)=\left\{\begin{array}{cc}
1 & v=0 \\
0 & v \neq 0
\end{array}\right.
\end{array}
\end{equation}

\begin{equation}
\begin{array}{c}
q_{\text {near }}=\underset{h}{\arg } \min \left\|\boldsymbol{z}^{\mathrm{q}}-\boldsymbol{z}^{h}\right\|^{2} \\
h \in\{1,2, \ldots, \mathrm{Q}\} \quad \text { and } \quad h \neq q
\end{array}
\end{equation}

Where $\left\{\mathbf{z}^{q}, l^{q}\right\}_{q=1}^{Q}$ denotes the dataset with their labels and $\varphi(.)$ denotes the Kronecker delta function. From (1) and (2), it is understood that the separation index is normalized between “zero” and “one”. 

Remark 1: In this paper, all applied distance norms as $\|x\|^{2}$ are $ L_2 $ norm.

\section{Separation index maximizing at the first convolutional layer }

\begin{figure}[ht]
\includegraphics[width=\linewidth]{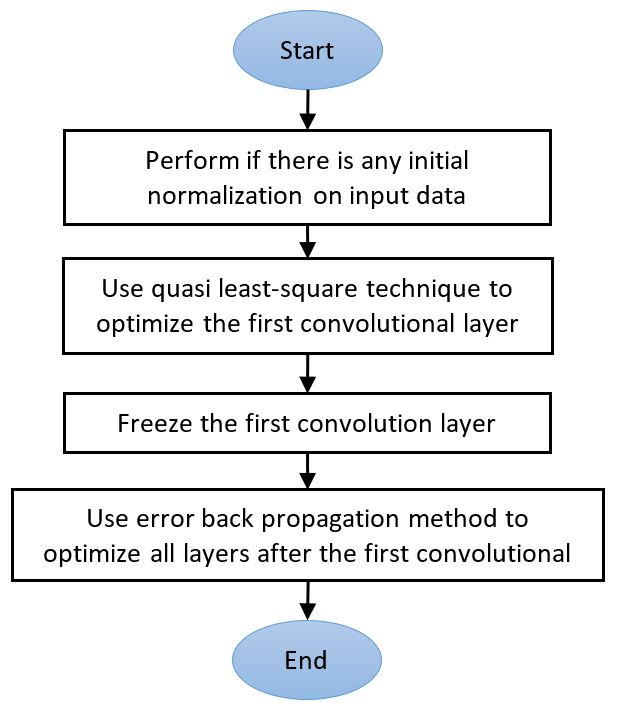}
\caption{This flowchart indicates the learning method in which the quasi-LS is utilized to learn the first convolution layer, and other layers are learned by an error backpropagation method.}
\label{fig:2}
\end{figure}

To increase the generalization of a DNN, it is required that SI increases along with the layers of a DNN, \cite{kalhor2019evaluation}. Using different error backpropagation methods indirectly cause increasing the SI. However, due to the vanishing gradient problem, the SI cannot significantly increase through initial layers, particularly at the first convolution layer as the base of the filtering process in a CNN. Here, to maximize the SI at the first layer of a CNN, quasi-LS technique is proposed, \cite{shults2014quasi}.  Fig \ref{fig:2} shows the flowchart which is used to learn a CNN, where only the first convolution layer is learned by quasi-LS, but the other layers are updated by the error backpropagation method.

\subsection{Quasi-Least Square Algorithm}

Assume there are Q input training patterns at input layer: $\left\{\boldsymbol{x}^{q}\right\}_{q=1}^{Q}$ Applying M convolutional filters and biases,$\left\{\left(\boldsymbol{\theta}_{l}, b_{l}\right)\right\}_{l=1}^{M}$ , to $\left\{\boldsymbol{x}^{q}\right\}_{q=1}^{Q}$ and then activating the units of M feature maps by Rectified Linear Unit (Relu), $\left\{\boldsymbol{Z}^{q}\right\}_{q=1}^{Q}$  are appeared, Fig \ref{fig:3}. In this paper, each $\boldsymbol{x}^{q}$ and $\boldsymbol{Z}^{q}$ are reshaped as a vector.

\begin{figure}[ht]
\includegraphics[width=\linewidth]{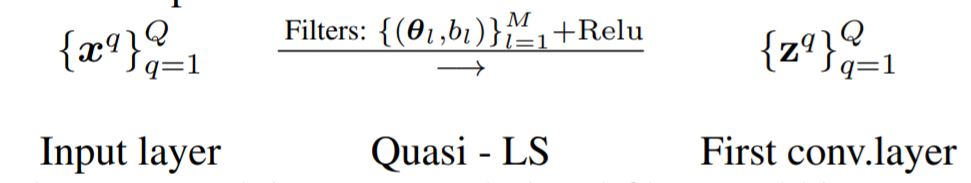}
\caption{Applying M convolutional filters and biases to the input patterns and then activating the convolved units by Relu function the feature maps are resulted. }
\label{fig:3}
\end{figure}
\raggedbottom

In order to maximize the SI at $\left\{z^{q}\right\}_{q=1}^{Q}$  ,it is aimed that at the first stage, filter parameters, $\left\{\left(\boldsymbol{\theta}_{l}, b_{l}\right)\right\}_{l=1}^{M}$ are initialized by using PCA technique, and then through some iterations, $\left\{\left(\boldsymbol{\theta}_{l}, b_{l}\right)\right\}_{l=1}^{M}$ are update in order that each  $\mathbf{z}^{q}$ becomes nearer to its nearest neighbor having the same label, $\boldsymbol{z}^{q_{f r}}$ ,and concurrently becomes far from its nearest neighbor having different label, $\boldsymbol{z}^{q_{e n}}$ . The considered goal will be satisfied by minimizing the following minimax loss function:
\raggedbottom
\begin{equation}
\begin{array}{c}
J\left(\left\{\left(\boldsymbol{\theta}_{l}, b_{l}\right)\right\}_{l=1}^{M}\right)= \\ \\
\sum_{q=1}^{Q} \gamma_{q}\left\|\mathbf{z}^{q}-\mathbf{z}^{q_{f r}}\right\|^{2}-\sum_{q=1}^{Q} \gamma_{q}\left\|\mathbf{z}^{q}-\mathbf{z}^{q_{e n}}\right\|^{2} \\ \\
\left\{\left(\boldsymbol{\theta}_{l}^{*}, b_{l}^{*}\right)\right\}_{l=1}^{M}=\arg \min J\left(\left\{\left(\boldsymbol{\theta}_{l}, b_{l}\right)\right\}_{l=1}^{M}\right)
\end{array}
\end{equation}

where $\gamma_{q}$ denotes the important weight of $\boldsymbol{z}^{q}$, which is explained later in the presented quasi-LS algorithm.
About the minimax loss function (3), it is required that the first part of the loss function is minimized and the second part is maximized. Ignoring the act of nonlinear activation function, the least-square (LS) technique is the best solution to minimize (3) but due to the nonlinearity and non-derivability of the activation function (Relu), gradient-based methods cannot be used for optimization purposes. However, the quasi-LS technique as a powerful hybrid optimization technique is employed. In a quasi-LS technique, at each iteration, by freezing the operation of the activation function, all positive units at the feature maps are participated in an LS technique to update the parameters of the filters. However, since after each updating, the operation of the activation function will change, the former LS technique should be repeated again. It is expected that by repeating the LS technique for several iterations, the filter parameters converge to more optimal parameters, \cite{shults2014quasi}. Table 2 presents the list of variables, which are used in the quai-LS algorithm.

\begin{table*}[h!]
\begin{center}
\begin{tabular}{|c|c|c|}
\hline
\textbf{Row} & \textbf{Variable} & \textbf{Explanation} \\
\hline\hline
1 &  $K$ & Number of classes in the classification problem\\
2 &  $Q$ & Number of all training patterns (each class has at least two patterns)\\
3 &  q & A counter variable for patterns $q \in\{1,2, \ldots, Q\}$ \\
4 &  $\boldsymbol{x}^{q}$ & The vector of all units in $q$th pattern at the input layer  \\
5 &  $\boldsymbol{z}^{q}$ & The vector of all units in $q$th pattern  at the first convolutional layer\\
6 &  $\boldsymbol{\gamma}^{q}$ & The important weight of $Z^{q}$ at each iteration.\\
7 &  $\mathbf{M}$ & The number of feature maps at the first convolutional layer.\\
8 & $n$  & The number of units at each feature map.\\
9 &  $l$ & A counter for number of units at each feature map. \\
10 & $n_{F}$  & The number of parameters at each filter without bias.\\
11 & $\boldsymbol{\theta}_{l} \in \mathbb{R}^{n_{F} \times 1}$ & the $l^{\text {th }}$ filter parameters.\\
12 &  $b_{l} \in \mathbb{R}$ & The $l^{th}$ bias corresponded to $l^{th}$ filter.\\
13 &  $\boldsymbol{\sigma}_{\rho}^{q} \in \mathbb{R}^{\mathbf{1} \times n_{F}}$ & The $\rho^{th}$ chosen patch from pattern $x_q$ .\\
14 &  $\boldsymbol{z}_{l}^{q} \in \mathbb{R}^{\boldsymbol{1} \times n}$ & The $l^{th}$ feature map from pattern $z_q$ .\\
15 & $z_{l \rho}^{q} \in \mathbb{R}$ & The $\rho^{th}$ unit of $l^{th}$ feature map from pattern $z_q$ .\\
16 &  $t$ & A counter for learning iterations in the Quasi-Least Square method.\\
17 & $t_{\text {stop }}$  & The number of iterations used in the Quasi-Least Square method.\\
\hline
\end{tabular}
\end{center}
\caption{The list of variables used in the quasi-least square method}
\end{table*}

\subsection{Quasi-LS algorithm}

The proposed quasi-LS algorithm is explained as follows:

\subsubsection{Stage1}
In this stage $\left\{\left(\widehat{\boldsymbol{\theta}}_{l}^{1}, \hat{b}_{l}^{1}\right)\right\}_{l=1}^{M}$ , as filter parameters and biases, are initialized by applying PCA on the input data: $\left\{x^{q}\right\}_{q=1}^{Q}$. It is assumed that M is an even number and $M \leq 2 n_{F}$. Considering that there are Q patterns at the first layer where each one has n patches, the mean of all patches of all the training patterns is computed as follows:

\begin{equation}
\overline{\boldsymbol{\sigma}}=\frac{1}{Q_{B}} \sum_{q=1}^{Q} \sum_{\rho=1}^{n} \boldsymbol{\sigma}_{p}^{q} \quad Q_{B}=n Q
\end{equation}

Now the covariance matrix of patches $\Sigma$ is defined as follow: 

$$
\Sigma=\frac{1}{Q_{B}} \tilde{X}^{T} \tilde{X} \quad \tilde{X}^{T}=\left[\widetilde{\sigma}_{1}^{1} \cdots \widetilde{\sigma}_{n}^{1} \widetilde{\sigma}_{1}^{2} \cdots \widetilde{\sigma}_{n}^{2} \cdots \widetilde{\sigma}_{1}^{Q} \cdots \widetilde{\sigma}_{n}^{Q}\right]
$$

\begin{equation}
\sigma_{p}^{q}=\sigma_{p}^{q}-\bar{\sigma}
\end{equation}

where $\sigma_{p}^{q}$ denotes the vector form of $p$ th patch of $q$ th example.
It is known that $\Sigma$ is a symmetric and positive semi-definite matrix. By applying singular value decomposition, to $\Sigma$:

\begin{equation}
\begin{array}{c}
\Sigma=\mathrm{V}^{\mathrm{T}} \Lambda \mathrm{V} \quad \mathrm{V}=\left[v_{1} v_{2} \ldots v_{n_{F}}\right] \\
\Lambda=\operatorname{diag}\left(\lambda_{1} \lambda_{2} \cdots \lambda_{n_{F}}\right) \lambda_{1} \geq \lambda_{2} \geq \cdots \geq \lambda_{n_{F}}
\end{array}
\end{equation}

where V as matrix of eigenvectors includes unitary orthogonal columns and $\Lambda$  as a diagonal matrix includes non-negative eigenvalues. Each eigenvector $v_{\tau}, \tau \in\left\{1,2, \ldots, n_{F}\right\}$ is corresponded with $\lambda_{\tau}$ and eigenvectors with bigger eigenvalues are more important because more data points are propagated along their determined directions. Now, all $M$ vectors of filter parameters and biases are initialized as follows: 

\begin{equation}
\begin{aligned}
\widehat{\boldsymbol{\theta}}_{l}^{1}=\left\{\begin{array}{ll}
\boldsymbol{v}_{\boldsymbol{\tau}} & \text { if } l=2 \tau-1 \\
-\boldsymbol{v}_{\boldsymbol{\tau}} & \text { if } l=2 \tau
\end{array} \quad \widehat{b}_{l}^{1}=-\overline{\boldsymbol{\sigma}}^{T} \widehat{\boldsymbol{\theta}}_{l}^{1}\right.\\
\tau \in\left\{1,2, \ldots, n_{F}\right\} \quad l \in\{1,2, \ldots, M\}
\end{aligned}
\end{equation}

\textbf{Remark 2 }: For when $M>2n_F$, one can initialize the first $2n_F$ filters by (7) but other $M-2n_F$ filters can be initialized by using a random initialization method.

\subsubsection{Stage2} Repeat following steps for t=1 up to $t=t_{\text {stop }}$:

\textit{Step1}: Define the matrix of patches, X, from patterns in the first layer:

\begin{equation}
\mathrm{X}=\left[\begin{array}{c}
\mathrm{X}^{1} \\
\vdots \\
\mathrm{X}^{Q}
\end{array}\right] \quad \mathrm{X}^{\mathrm{q}}=\left[\begin{array}{cc}
\boldsymbol{\sigma}_{1}^{q} & 1 \\
\vdots & \vdots \\
\boldsymbol{\sigma}_{n}^{q} & 1
\end{array}\right]
\end{equation}

\textit{Step2}: Update the patterns at layer L as follows:

\begin{equation}
\begin{array}{c}
\widehat{\boldsymbol{z}}^{q}=\left[\hat{\boldsymbol{z}}_{1}^{q} \hat{\boldsymbol{z}}_{2}^{q} \cdots \hat{\boldsymbol{z}}_{\boldsymbol{M}}^{\boldsymbol{q}}\right] \\
\hat{\boldsymbol{z}}_{l}^{q}=\left[\hat{z}_{l, 1}^{q} \hat{z}_{l, 2}^{q} \ldots \ldots . \hat{z}_{l, n}^{q}\right] \\
\hat{z}_{l, \rho}^{q}=\operatorname{Relu}\left(\boldsymbol{\sigma}_{p}^{q} \widehat{\boldsymbol{\theta}}_{l}^{t}+\widehat{b}_{l}^{t}\right)
\end{array}
\end{equation}

\textit{Step3}: For $\hat{\boldsymbol{z}}^{q}$, find the nearest neighbor pattern with the same label (friend pattern) as follows:

\begin{equation}
\widehat{\boldsymbol{z}}^{q_{f r}}=\operatorname{near}_{f r}\left(\hat{\boldsymbol{z}}^{q}\right)
\end{equation}

where $\operatorname{near}_{f r}\left(\hat{\mathbf{z}}^{q}\right)$ denotes a function which finds the nearest neighbor pattern from patterns which have the same label with $\widehat{\boldsymbol{z}}^{q}$

Now, decompose $\widehat{\boldsymbol{z}}^{q_{f r}}$ as follows:

\begin{equation}
\begin{array}{l}
\hat{\boldsymbol{z}}^{q_{f r}} \rightarrow \hat{\boldsymbol{z}}^{q_{f r}}=\left[\hat{\boldsymbol{z}}_{1}^{q_{f r}} \hat{\boldsymbol{z}}_{2}^{q_{f r}} \cdots \hat{\boldsymbol{z}}_{M}^{q_{f r}}\right] \\
\widehat{\boldsymbol{z}}_{l}^{q_{f r}} \rightarrow \widehat{\boldsymbol{z}}_{l}^{q_{f r}}=\left[\hat{z}_{l, 1}^{q_{f r}} \hat{z}_{l, 2}^{q_{f r}} \cdots \hat{z}_{l, n}^{q_{f r}}\right]
\end{array}
\end{equation}

\textit{Step4}: For $\widehat{\mathbf{z}}_{q}^{L}$, find the nearest neighbor pattern from patterns which have different label with $\widehat{\boldsymbol{z}}^{q}$ :

\begin{equation}
\hat{\boldsymbol{z}}^{q_{e n}}=\operatorname{near}_{e n}\left(\hat{\boldsymbol{z}}^{q}\right)
\end{equation}

where $\operatorname{near}_{e n}\left(\hat{\mathbf{z}}^{q}\right)$ denotes a function which finds the nearest neighbor pattern from patterns which have different label with $\widehat{\boldsymbol{z}}^{q}$

Now, decompose $\widehat{\boldsymbol{Z}}^{q_{e n}}$ as follows:

\begin{equation}
\begin{array}{l}
\hat{\mathbf{z}}^{q_{e n}} \rightarrow \hat{\mathbf{z}}^{q_{e n}}=\left[\hat{\boldsymbol{z}}_{1}^{q_{e n}} \hat{\mathbf{z}}_{2}^{q_{e n}} \cdots \hat{\mathbf{z}}_{M}^{q_{e n}}\right] \\
\widehat{\mathbf{z}}_{l}^{q, e n} \rightarrow \widehat{\mathbf{z}}_{l}^{q_{e n}}=\left[\hat{z}_{l, 1}^{q_{e n}} \hat{z}_{l, 2}^{q_{e n}} \cdots \hat{z}_{l, n}^{q_{e n}}\right]
\end{array}
\end{equation}

\textit{Step5}: Form following data matrices including     
\begin{equation}
Z_{l}^{f r}=\left[\begin{array}{c}
Z_{l}^{1} f r \\
\vdots \\
Z_{l}^{Q_{f r}}
\end{array}\right] \quad Z_{l}^{q_{f r}}=\left[\begin{array}{c}
z_{l, 1}^{q_{f r}} \\
\vdots \\
z_{l, n}
\end{array}\right]
\end{equation}

\begin{equation}
Z_{l}^{e n}=\left[\begin{array}{c}
Z^{1} e n \\
\vdots \\
Z^{Q_{e n}}
\end{array}\right] \quad Z_{l}^{q_{e n}}=\left[\begin{array}{c}
z_{l, 1}^{q_{e n}} \\
\vdots \\
z_{l, n}^{q_{e n}}
\end{array}\right]
\end{equation}

\begin{figure*}[ht]
\includegraphics[width=\linewidth]{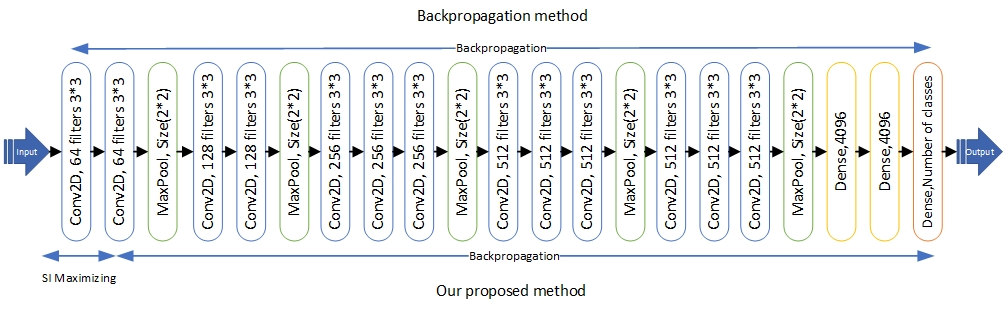}
\caption{An example of the difference between implementing our method and the backpropagation method is shown in this image for vgg16.}
\label{fig:4}
\end{figure*}

\textit{step6}: define the weight matrix of all patches of all patterns as follows:

\begin{equation}
\begin{array}{c}
\Phi_{l}=\left[\begin{array}{ccc}
\Phi_{l}^{1} & \mathbf{0} & 0 \\
\mathbf{0} & \ddots & \mathbf{0} \\
0 & \mathbf{0} & \Phi_{l}^{Q}
\end{array} \right] \\ \\
\Phi_{l}^{q}=\gamma_{q}\left[\begin{array}{ccc}
\operatorname{sign}\left(\hat{z}_{l, 1}^{q}\right) & \mathbf{0} & 0  \\
\mathbf{0} & \ddots & \mathbf{0} \\
0 & \mathbf{0} & \operatorname{sign}\left(\hat{z}_{l, n}^{q}\right)
\end{array}\right]
\end{array}
\end{equation}

where $\gamma_{q}$ denotes the important weight of  $q t h$ pattern. $\gamma_{q}$ is defined as follows:

\begin{equation}
\gamma_{q}=\left\{\begin{array}{cc}
2 & \text { if }\left\|\hat{\boldsymbol{Z}}^{q}-\hat{\boldsymbol{z}}^{q_{e n}}\right\| \leq\left\|\hat{\mathbf{z}}^{q}-\hat{\boldsymbol{z}}^{q_{f r}}\right\| \\
1 & \text { otherwise }
\end{array}\right.
\end{equation}

With regard to (17), when the nearest neighbor of $\hat{\boldsymbol{z}}^{q}$q does not have the same label with it, the important weight of $\hat{\boldsymbol{z}}^{q}$ increases to 2, in quasi-LS technique. 

\textit{Step7}: update the parameters of filters and biases. At first, compute following vector:

\begin{equation}
\begin{array}{c}
\boldsymbol{\psi}^{t}=\boldsymbol{\psi}_{f r}^{*}-\boldsymbol{\psi}_{e n}^{*} \quad \boldsymbol{\psi}_{\mathrm{fr}}^{*}=\underset{\boldsymbol{\psi}_{\mathrm{fr}}}{\arg \min }\left\|\mathrm{X} \boldsymbol{\psi}_{\mathrm{fr}}-Z_{l}^{f r}\right\|^{2} \\
\boldsymbol{\psi}_{e n}^{*}=\underset{\boldsymbol{\psi}_{e n}}{\arg \min \| \mathrm{X} \boldsymbol{\psi}_{e n}}-Z_{l}^{e n} \|^{2}
\end{array}
\end{equation}

where $\boldsymbol{\psi}_{f r}^{*}$ and $\boldsymbol{\psi}_{e n}^{*}$ are optimized by least-square technique as follows:

\begin{equation}
\begin{array}{l}
\boldsymbol{\psi}_{\mathrm{fr}}^{*}=\left(\mathrm{X}^{\mathrm{T}} \Phi_{1}^{\mathrm{t}} \mathrm{X}\right)^{-\mathbf{1}} \mathrm{X}^{\mathrm{T}} \Phi_{l} Z_{l}^{f r} \\
\boldsymbol{\psi}_{e n}^{*}=\left(\mathrm{X}^{\mathrm{T}} \Phi_{1}^{\mathrm{t}} \mathrm{X}\right)^{-\mathbf{1}} \mathrm{X}^{\mathrm{T}} \Phi_{l} Z_{l}^{e n}
\end{array}
\end{equation}

Now, update the parameters of filters and biases as follows:

\begin{equation}
\left[\begin{array}{l}
\widehat{\boldsymbol{\theta}}_{l}^{t+1} \\
\hat{b}_{l}^{t+1}
\end{array}\right]=\left[\begin{array}{l}
\widehat{\boldsymbol{\theta}}_{l}^{t} \\
\hat{b}_{l}^{t}
\end{array}\right]+\boldsymbol{\psi}^{t}
\end{equation}

\subsubsection{Satge3}
Put $\boldsymbol{\theta}_{l}=\widehat{\boldsymbol{\theta}}_{l}^{t-1}$ and $b_{l}=\hat{b}_{l}^{t-1}$ as final opted filter parameters and biases at the first convolutional layer.

\begin{table}[t]
\begin{center}
\begin{tabular}{|c|c|c|}
\hline
\textbf{Parameter}                                 & \multicolumn{2}{c|}{\textbf{Value}}                     \\ \hline \hline
Optimizer                                 & \multicolumn{2}{c|}{Adam}                      \\ \hline
Preformance metric                        & \multicolumn{2}{c|}{Accuracy}          \\ \hline
Loss function                             & \multicolumn{2}{c|}{Categorical Cross-Entropy} \\ \hline
Batch Size                                & 
\multicolumn{2}{c|}{128}                        \\ \hline
Learning Rate                                & 
\multicolumn{2}{c|}{0.01}                        \\ \hline
Epoch                               & 
\multicolumn{2}{c|}{250}                        \\ \hline
\end{tabular}
\end{center}
\caption{Simulation parameters of the image classification networks (These parameters are valid for both the Backpropagation method and the Backpropagation section of our proposed method)}
\end{table}


\begin{table}[ht]
\begin{center}
\begin{tabular}{|c|c|c|c|}
\hline
Feature/Dataset & CIFAR10  & CIFAR100  & F-MNIST  \\ \hline
\# of train     & 50000   & 50000    & 60000         \\ \hline
\# of test      & 10000   & 10000    & 10000         \\ \hline
\# of classes   & 10      & 100      & 10            \\ \hline
color mode      & RGB     & RGB      & GRAY          \\ \hline
dimensions      & 32*32*3 & 32*32*3  & 28*28*1       \\ \hline
\end{tabular}
\end{center}
\caption{Datasets specification used in the paper}
\end{table}

\section{Experimental Results}

\begin{table*}[t]
\begin{center}
\begin{tabular}{|c|l|c|c|c|c|c|}
\hline  

\multicolumn{2}{|c|}{Dataset}     &                  Learning Method       & \textbf{AlexNet } & \textbf{VGG16} & \textbf{ResNet50} & \textbf{InceptionV3} \\ \hline \hline

\multicolumn{2}{|c|}{CIFAR10}       & Backpropagation         & 84.61    & 92.95  & 93.17     & 94.20        \\ \cline{3-7} 
\multicolumn{2}{|c|}{}                               & Our proposed method     & \textbf{85.84}    & \textbf{94.81}  & \textbf{95.02}  & \textbf{95.43}       \\ \cline{3-7} 
\multicolumn{2}{|c|}{}                               & Percentage of improvement     & 1.23    & 1.86  & 1.85  & 1.23       \\ \hline
\multicolumn{2}{|c|}{CIFAR10 - (plane,truck)}       & Backpropagation         & 96.45    & 97.18  & 97.64     & 97.09        \\ \cline{3-7} 
\multicolumn{2}{|c|}{}                               & Our proposed method     &\textbf{ 97.09  }  & \textbf{98.36}  & \textbf{98.49}  & \textbf{97.77}       \\ \cline{3-7} 
\multicolumn{2}{|c|}{}                               & Percentage of improvement     & 0.64    & 1.18  & 0.85  & 0.68       \\ \hline
\multicolumn{2}{|c|}{CIFAR10 - (plane,cat,bird)}       & Backpropagation         &  96.21    & 96.80  & 96.88  & 96.81        \\ \cline{3-7} 
\multicolumn{2}{|c|}{}                               & Our proposed method   &  \textbf{96.62}    & \textbf{97.63}  & \textbf{97.16}  & \textbf{97.14}       \\ \cline{3-7} 
\multicolumn{2}{|c|}{}                               & Percentage of improvement    &  0.41   & 0.83  & 0.28  & 0.33       \\ \hline

\multicolumn{2}{|c|}{CIFAR100}      & Backpropagation         & 62.22   & 70.98  & 75.30   & 76.31        \\ \cline{3-7} 
\multicolumn{2}{|c|}{}                               & Our proposed method     & \textbf{62.45}    & \textbf{71.74}  & \textbf{75.58}     & \textbf{76.72}        \\ \cline{3-7} 
\multicolumn{2}{|c|}{}                               & Percentage of improvement     & 0.23   & 0.76  & 0.28  & 0.41       \\ \hline

\multicolumn{2}{|c|}{ASHION-MNIST} & Backpropagation         & 92.53    & 94.17  & 95.24     & 95.78      \\ \cline{3-7} 
\multicolumn{2}{|c|}{}                               & Our proposed method     & \textbf{92.65 }   & \textbf{94.73}  & \textbf{95.38}    & \textbf{95.91}        \\  \cline{3-7} 
\multicolumn{2}{|c|}{}                               & Percentage of improvement     & 0.12   & 0.56  & 0.14  & 0.13       \\ \hline

\end{tabular}
\end{center}
\caption{Comparing the accuracy of our proposed method in different architectures and datasets with backpropagation method}
\end{table*}

\subsection{Implementation Details}

In our proposed method, we first increase the SI value in the first layer, then train the network in the other layers by the backpropagation method. However, the weights of the first layer are freeze and do not change.

To evaluate the proposed method, our proposed method and backpropagation method has been evaluated on four different architectures and four different datasets. Four state-of-the-art convolutional neural network architectures, including AlexNet and VGG16, ResNet50, InceptionV3 are used. In all four architectures, the weights of the first layer are learned by our proposed method to achieve the highest possible SI in the first layer, then for the next layer training to end layer, the backpropagation method is used. An example of the proposed method is shown in Fig \ref{fig:4} for the VGG16 architecture.

In the backpropagation method, the parameters of Table 3 are used. Also, for the train of the backpropagation section of our proposed method, the parameters of Table 3 are used exactly, so that the conditions for a fair comparison of the two methods are provided. Also, for each architecture, dataset, and method, the network is trained 4 times and the average accuracy obtained these 4 times is calculated and placed in Table 4. The reason for this is that the result obtained with no chance.

\subsection{Datasets \& Architectures}

In this paper, 4 state-of-the-art convolutional neural network architectures including AlexNet \cite{krizhevsky2012imagenet}, VGG16 \cite{simonyan2014very}, InceptionV3\cite{szegedy2016rethinking}, and ResNet50\cite{he2016deep}, and 3 state-of-the-art datasets including CIFAR10 \cite{krizhevsky2009learning}, CIFAR100\cite{krizhevsky2009learning}, and FASHION-MNIST \cite {xiao2017fashion} (F-MNIST) were used to evaluate the proposed method.

\subsection{Evaluation Metric}

Different metrics are used to evaluate network performance and comparison with other networks. In this paper, we use accuracy for comparison between our proposed method and the backpropagation method. How to calculate the accuracy is specified in (21)

\begin{equation}
Accuracy =    \frac{TP  + TN}{P + N} =  \frac{TP + TN}{TP + TN + FP + FN} 
\end{equation}

\subsection{Result}

The results of the experiments are given in Table 4, as shown in Table 4, in most architectures and datasets the accuracy be increased by our proposed method. This improvement is greater in datasets with fewer classes, as shown in Table 5, The CIFAR10 dataset has the same number of attributes as the CIFAR100 dataset, but the CIFAR10 dataset has fewer classes. And the performance of our proposed method in the CIFAR10 dataset is better than the CIFAR100 dataset. Also, in the dataset with more features, a higher accuracy increase is obtained, as shown in Table 4, the increase in accuracy of CIFAR10 and CIFAR100 was more than FASHION-MNIST. Also in the CIFAR10 dataset in 2-class mode, the accuracy increase ratio is more than 3-class mode.

\section{Conclusion}

This paper presents a separation index-based method that acts as a preprocessor in the first layer of convolutional neural networks. Using the proposed method, we increase the value of the separation index in this first layer as much as possible, and then by Using backpropagation, the rest of the layers are trained. The proposed method and backpropagation method were trained four times on GPU for each architecture and dataset. Finally, the average accuracy obtained from both methods was compared, which shows improved accuracy with our proposed method compared to the backpropagation method.

Among the things that can be done in the following of this article is to do more experiments of the proposed method on more architecture and datasets, as well as the use of separation index in more layers or even all layers for network training.

\bibliography{mybibfile}

\end{document}